\documentclass{article}
\usepackage{spconf,amsmath,amssymb,graphicx}

\title{Using recurrences in time and frequency within U-net architecture for speech enhancement}
\twoauthors
  {Tomasz Grzywalski}
	{StethoMe\textsuperscript{\textregistered}\\
	Poznan\\
	Poland}
  {Szymon Drgas} 
	{Institute of Automation and Robotics\\
	Poznan University of Technology\\
	Poland}
\begin{document}
%



\maketitle

\begin{abstract}
When designing fully-convolutional neural network, there is a trade-off between receptive field size, number of parameters and spatial resolution of features in deeper layers of the network. In this work we present a novel network design based on combination of many convolutional and recurrent layers that solves these dilemmas. We compare our solution with U-nets based models known from the literature and other baseline models on speech enhancement task. We test our solution on TIMIT speech utterances combined with noise segments extracted from NOISEX-92 database and show clear advantage of proposed solution in terms of SDR (signal-to-distortion ratio), SIR (signal-to-interference ratio) and STOI (spectro-temporal objective intelligibility) metrics compared to the current state-of-the-art.
\end{abstract}

\section{Introduction}
{\let\thefootnote\relax\footnotetext{This work has been submitted to the IEEE for possible publication. Copyright may be transferred without notice, after which this version may no longer be accessible.}}

The single-channel speech enhancement problem is to reduce a noise present in a single-channel recording of speech. This technique has many applications, it can be employed as a preprocessing step in speech or speaker recognition system \cite{zhang2018deep} or to improve speech intelligibility what is important for example in hearing aids like cochlear implants \cite{lai2018deep}. 

Recently, data-driven approaches became popular for speech enhancement. In these methods training data is used to train model whose aim is to reduce even nonstationary noise. Data-driven methods include methods based on non-negative matrix factorization \cite{fevotte2018single} and deep neural networks (DNNs)\cite{wang2018supervised}. 
DNNs are used as a nonlinear transformation of a noisy signal to a clean one (mapping-based targets), or to a filtering mask (i.e. time-varying filter in the short-time Fourier transform domain), that can be used to recover speech (masking-based targets).

In recent work \cite{wang2018supervised}  deep learning methods for speech enhancement are summarized. In the early DNN-based methods for speech enhancement, neural network with fully-connected layers acted as a mapping from a spectrogram fragment of a noisy speech (a given spectrogram frame and its context) to a spectrogram frame of enhanced speech \cite{huang2014deep,huang2015joint}. 

Convolutional neural networks (CNNs) were applied to speech enhancement in \cite{7394335}, by combining convolutional and fully connected layers to estimate the ideal binary mask (IBM). 
In \cite{grais2017single} fully convolutional network was used. In this case encoder-decoder architecture was employed, where a number of convolutive layers interleaved by max-pooling act as encoder and the same number of convolutive layers, but interleaved by upsampling act as decoder. Decoder maps activations (feature map) at the output of the encoder to the magnitude spectrum of enhanced speech. In this case however, separation quality may be limited as max-pooling operation, which is irreversible, reduces time-frequency resolution of subsequent feature maps. 

The similar problem in the domain of medical image segmentation was mitigated in U-net architecture \cite{ronneberger2015u}, by using skip connections. In \cite{jansson2017singing} U-net architecture brought better singing voice separation quality in comparison to a network without skip connections. In convolutional neural networks used in \cite{park2016fully} an architecture similar to U-net was proposed, and the authors showed that removal of max-pooling and upsampling may be beneficial in terms of separation quality.

In \cite{zhao2018convolutional} recurrent-convolutional architecture was proposed for speech enhancement. It consists of convolutional layers followed by bidirectional recurrent component, finally fully-connected layer is used to predict the clean spectrogram. The authors obtained improved performance in comparison to DNN with fully-connected layers only and recurrent neural networks.

In \cite{grzywalski2018application}  recurrent U-net architecture for speech enhancement was proposed. The results suggested that max-pooling in U-net introduces loss of information on deeper levels, but it is needed to build big enough receptive field (context in the input spectrogram on which the corresponding element of the output spectrogram depends). However recurrent layers can enlarge receptive field so that max-pooling is no longer needed. On the average the best combination was to build a network without max-pooling and to use recurrent layers to extend the receptive field.

In \cite{visin2015renet} ReNet architecture was proposed which replaces convolution and pooling layers with four recurrent layers that sweep horizontally and vertically in both directions across the image. The use of horizontal and vertical layers alternately better scales with number of dimensions in comparison to multidimensional RNNs (recurrent neural networks) \cite{graves2009offline}. Evaluation performed on image data suggest that ReNet gives comparable results to CNNs.

In this work we further develop idea presented in \cite{grzywalski2018application}. We introduce recurrent-convolutional (RC) pairs which are the blocks from which the proposed network is built. Each output of RC pair can potentially depend on a bigger context of its input than convolutional layer. This is for the cost of relatively small number of additional parameters. Moreover, this context can be enhanced at many depths of the encoder and decoder. 


\section{Description of the proposed architecture}
\label{sec:proposed}
\begin{figure}
    \centering
    \includegraphics[width=8.4cm]{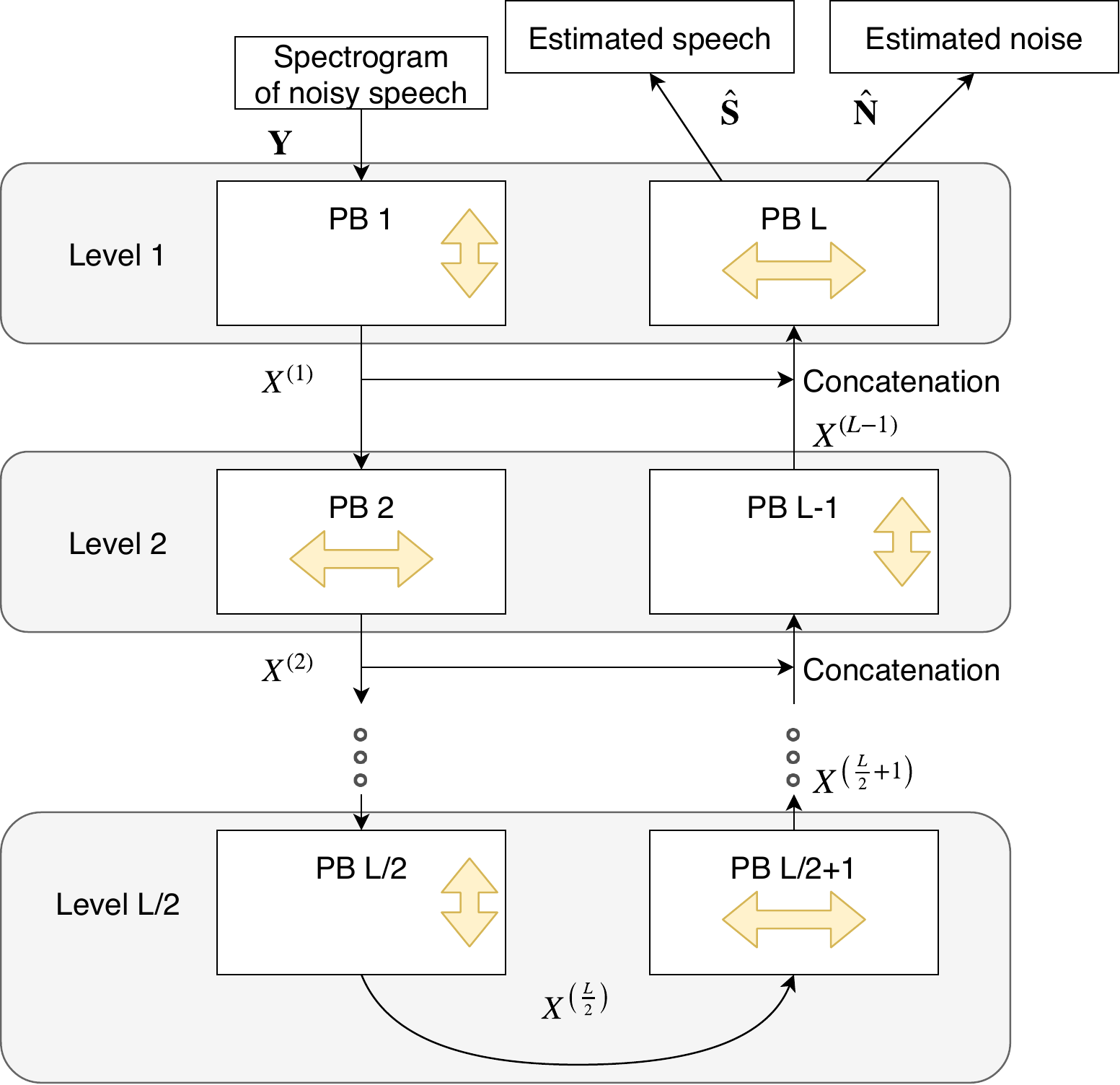}
    \caption{General architecture of the proposed U-nets}
    \label{fig:unet}
\end{figure}
The proposed neural network represents a function \linebreak ${(\hat{\bf S},\hat{\bf N})=f({\bf Y};\Theta)}$, where ${\bf Y}\in\mathbb{R}^{B\times N}$ denotes spectrogram of a noisy speech, $B$ denotes the number of frequency channels, while $N$ is the number of spectrogram frames. The neural network is trained to obtain at the output  matrices $\hat{\bf S}\in\mathbb{R}^{B\times N}$ and $\hat{\bf N}\in\mathbb{R}^{B\times N}$ representing magnitude spectrograms of a clean speech and noise respectively. 

The general structure of the proposed neural network architecture is shown in Figure \ref{fig:unet}. The network consists of processing blocks PB 1, \ldots, PB L. 
Each processing block PB~$l$, where $l=1,\ldots,L$ accepts a feature map (tensor) of dimension $B \times N \times K^{(l)}_{\text{in}}$ and outputs feature map \linebreak $X^{(l)}\in\mathbb{R}^{B\times N\times K^{(l)}_{\text{out}}}$, where $K_{\text{in}}^{(l)}$ and $K_{\text{out}}^{(l)}$ are the numbers of features in the input and output feature maps respectively. The input to PB 1, is the input spectrogram ${\bf Y}$ represented as a tensor  with $K_{\text{in}}^{(1)}=1$, the inputs to PB $l$ for $l=2,\ldots,L/2+1$ are $X^{(l-1)}$, while for $l=L/2+2,\ldots,L$ the input is a concatenation of $X^{(l-1)}$ and the output feature map of processing block from the encoder at the same level. The concatenation is done along the third dimension of the feature maps.

In the original U-net \cite{ronneberger2015u} each processing block comprises convolutional layers, max-pooling or upsampling operations. In this work we propose to use RC pairs described in the subsequent sections. In RC pairs recurrences are used to enlarge receptive field in time or frequency dimension. In proposed solution we use RC pairs with alternate dimensions in consecutive processing blocks. An example is shown in Figure \ref{fig:unet} where the bidirectional arrows show the dimension in which the context is enhanced (vertical and horizontal arrows correspond to frequency and time dimensions respectively). In this configuration for each path between input and output, blocks with enhanced context in time dimensions are interleaved with blocks with enhanced context in frequency dimension.

\subsection{Recurrent-convolutional (RC) pairs}
The recurrent-convolutional (RC) pair is shown in Figure \ref{fig:bwrl_rc} (left). Let $I\in \mathbb{R}^{B\times N\times K_{\text{in}}}$ be an input feature map to the block RC. In comparison to a processing block with a convolutional layer only (with nonlinearity), in the RC pair, map $I$ is concatenated with additional features extracted by the recurrent (BWR) layer $R\in\mathbb{R}^{B\times N\times K_{\text{rec}}}$ which results in \linebreak $C\in\mathbb{R}^{B\times N\times (K_{\text{in}}+K_{\text{rec}})}$.

Because of the skip connection, even when the number of features extracted by BWR ($K_{\text{rec}}$) is small, the context in $I$ on which each feature in the output feature map $O$ depends can be substantially increased.

\subsection{Bidirectional weight-sharing recurrences}
\begin{figure}
    \centering
    \includegraphics[width=8.4cm]{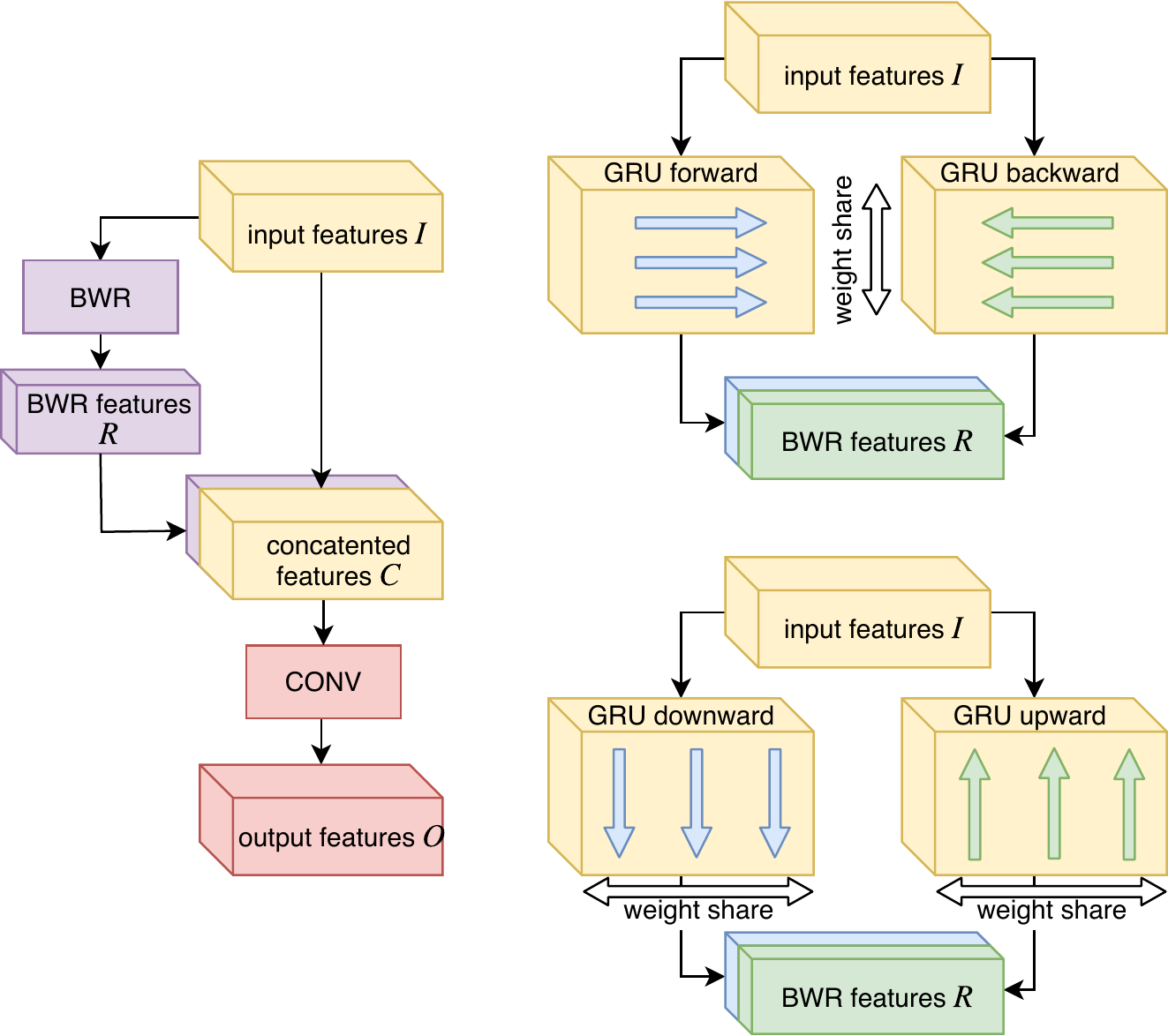}
    \caption{Left: block diagram of RC pair; right: block diagrams of two types of BWR (top: BWR\textsubscript{T} bottom: BWR\textsubscript{F})}
    \label{fig:bwrl_rc}
\end{figure}

BWR layers are shown in Figure \ref{fig:bwrl_rc} (right). They consists of a pair of recurrent layers iterating in opposite directions, whose outputs are concatenated along the third dimension. There are two types of BWR layers: time (T) and frequency (F) depending on the spatial dimension in which they iterate over input feature map $I$. BWR\textsubscript{T} accepts on input a sequence of $N$ $K_{\text{in}}$-dimensional vectors representing single frequency band $b$. The same weights are used to iterate over all $B$ frequency bands. BWR\textsubscript{F} accepts on input a sequence of $B$ $K_{\text{in}}$-dimensional vectors representing single time frame $n$. The same weights are shared to iterate over all $N$ time frames.

\section{Experiments}
\label{sec:experiments}

The noisy speech examples were obtained by mixing TIMIT \cite{timit} speech utterances with noise segments extracted from NOISEX-92 database. The training and test datasets contain 2000 and 192  utterances respectively. Both speech and noise signals were resampled to 8000 Hz. The speech enhancement quality was assessed for babble and factory noises, mixed with the speech utterances at SNR 0 dB. 
\subsection{Feature extraction}
For all audio signals short-time Fourier transform were computed. The frame step was 10 ms, while frame length was 25 ms. Hann window was applied. For each frame 512-point FFT was calculated. Afterwards, 64-channel mel-scale filterbank was applied to the magnitude of STFT. Finally, element-wise logarithm was calculated from the resulting STFT matrices.

\subsection{Baseline architectures}
We compared our proposed solution with two baseline models defined in \cite{chen2017long}. Each model was optimized to get best performance on our dataset.

Fully connected layers network (FCLN) accepts the same spectrogram on its input as the proposed U-net architectures and uses sliding-window technique with window length of 23 frames (11 to the left and 11 to the right). The output of the network is IRM (ideal ratio mask) \cite{wang2014training} mask. Optimized version of this network featured 4 hidden layers with 512 neurons each. Additionally we used ELU nonlinearities instead of RELU and linear outputs instead of sigmoid for the output layer which also helped to improve separation quality.

Recurrent neural network (RNN) uses the same sliding-window scheme as FCLN network. RNN originally consisted of 4 hidden layers with 512 long short-term memory (LSTM) units. We found these numbers optimal, but for each forward recurrence we added a second one going in the opposite direction, effectively making it a bidirectional layer with 1024 units, what improved separation quality. We also added gradient clipping above 100 and, as in FCLN baseline, removed sigmoid nonlinearities on output. Similar to FCLN, this network also predicts IRM mask.

We also considered training baseline models (FCLN and RNN) without IRM (for direct estimation of clean speech and noise like in our proposed solution), but the initial experiments have shown very low performance so we decided not not investigate this scenario further.

\subsection{U-nets}
\begin{figure*}[!t]
    \centering
    \includegraphics[width=7.0in]{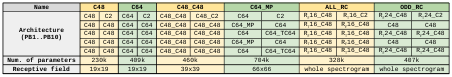}
    \caption{Evaluated U-nets, first four are reference baseline U-nets, last two implement our proposed solution; architectures are described by defining each processing block from Figure \ref{fig:unet} in form of a 5x2 table (left column - encoder, right - decoder)}
    \label{fig:netarch}
\end{figure*}

 Based on scheme presented in Figure \ref{fig:unet} we define four baseline U-net architectures and two that implement RC pairs. All networks feature 5 levels ($L = 10$ processing blocks). Figure \ref{fig:netarch} shows architectures of all six networks along with number of parameters and receptive field sizes.

\noindent We use following notation:

\noindent $\bf{C48}$: 2D convolutional layer with 48 filters (all convolutional layers use 3x3 filters with ELU and batch normalization except for final layer which is always 1x1 with linear output),

\noindent $\bf{R\textsubscript{T}16\_C48}$: RC pair comprising BWR layer with 8 units per direction, iterating in time axis (weight sharing in frequency axis), and C48 layer (all recurrences were implemented using Gated Recurrent Units (GRUs) with gradient clipping above 100 and batch normalization),

\noindent $\bf{R\textsubscript{F}16\_C48}$: frequency counterpart to R\textsubscript{T}16\_C48,

\noindent ${\bf MP}$: max-pooling 2x2,

\noindent ${\bf TC48}$: transposed convolution with filter size 6x6, stride 2 and crop 2 (an inverse of standard 3x3 convolution with "same" padding followed by 2x2 max-pooling).

We also tested ALL\_RC model for predicting IRM, this allowed for more direct comparison with the baseline models. In this scenario final layer consisted of single 1x1 filter with linear output.

\subsection{Reconstruction and metrics}
The magnitude mel-spectrogram in log-scale of the clean speech was estimated using the proposed neural networks. After applying exponential function, the mel-filtering was inverted by means of the pseudoinverse of the matrix with characteristics of the mel-filters. Next, the result was combined with phase of the noisy speech. This allowed to reconstruct the speech signal.

In order to assess the quality of the separation for different variants of the proposed architecture, SDR (signal-to-distortion ratio), SIR (signal-to-interference ratio), and SAR (signal-to-artifact) ratio were implemented as defined in \cite{vincent2006performance}. Additionally we used spectro-temporal objective intelligibility (STOI) as defined in \cite{taal2011algorithm}.

\subsection{Meta parameters}
All networks were trained for 100 epochs using Adam optimizer with batch size of 15. For all networks initial learning rate was set to 0.001 except for networks with RC pairs where learning late 0.01 was used. Higher learning rate slightly improved results for these networks, no such improvement was observed for the remaining networks. Learning rate was multiplied by 0.99 after every epoch.

All presented U-nets (except for ALL\_RC IRM experiment) were trained to minimize absolute difference between actual and predicted spectrograms of clean speech and noise.

10\% of training data was held out as validation set. Best model snapshot was selected based on SDR obtained on validation set. Validation was performed after each epoch.

\subsection{Results}
The results of the performed experiments are shown in tables \ref{tab:res_factory} and \ref{tab:res_babble}. The proposed architecture with RC pairs gave the best separation quality in terms of all metrics. ALL\_RC and ODD\_RC gave the same SDRs and STOIs for factory noise, for the babble noise these two architectures also had comparable results (difference on SDR and STOI was 0.1 dB and 0.01 respectively). ALL\_RC brought higher SIR than ODD\_RC by 0.5 dB and 0.9 dB for babble and factory noise respectively. It was, however, slightly worse in comparison to ODD\_RC in terms of SAR.

It can be noticed that U-net architectures without recurrent layers, provide higher SDRs and SIRs in comparison to the baseline models. This is not the case for SAR and STOI. However, U-nets with RC pairs give improvement in terms of SDR, SIR, and STOI. In comparison to the best non-U-net baseline (RNN IRM), both ALL\_RC and ODD\_RC gave improvement of 0.9 dB of SDR and 0.05 of STOI for the factory noise. In the case of babble noise, this difference is bigger, i.e. 1.4 dB for SDR and 0.07 for STOI. The best architectures with RC pairs outperformed also U-net architectures without recurrences. For the factory noise the difference was 0.2 dB, while for the babble noise it was 0.7 dB. The improvement can be also observed for STOI metric -- 0.05 for factory and 0.06 for babble noise.

\begin{table}
\caption{Factory noise}
\begin{tabular}{lllll}

\hline
 Name                       & SDR  & SIR   & SAR  & STOI \\
 \hline
 FCLN IRM                   & 7.4  & 12.8  & 8.9  & 0.74     \\
 RNN IRM                    & 7.5  & 12.2  & 9.3  & 0.76     \\
 \hline
 U-net C48                  & 7.9  & 14.7  & 8.9  & 0.73     \\
 U-net C64                  & 8.0  & 14.2  & 9.1  & 0.73     \\
 U-net C48\_C48             & 8.2  & 14.3  & 9.3  & 0.76     \\
 U-net C64\_MP              & 8.1  & 14.5  & 9.3  & 0.76     \\
 \hline
 U-net ALL\_RC IRM          & 8.2  & 14.8  & 9.3  & 0.80     \\
 U-net ALL\_RC              & {\bf 8.4} & {\bf 15.5} & 9.4 & {\bf 0.81} \\
 U-net ODD\_RC              & {\bf 8.4} & 15.0 & {\bf 9.5} & {\bf 0.81}    \\
\hline
\end{tabular}
\label{tab:res_factory}
\end{table}

\begin{table}
\caption{Babble noise}
\begin{tabular}{lllll}
\hline
 Name                       & SDR  & SIR   & SAR  & STOI \\ \hline
 FCLN IRM                   & 5.3  & 8.9   & 8.5  & 0.71     \\
 RNN IRM                    & 5.6  & 9.2   & 8.7  & 0.72     \\
 \hline
 U-net C48                  & 6.1  & 11.9  & 7.8  & 0.69     \\
 U-net C64                  & 6.1  & 11.6  & 7.8  & 0.69     \\
 U-net C48\_C48             & 6.2  & 10.5  & 8.7  & 0.71     \\
 U-net C64\_MP              & 6.3  & 11.2  & 8.4  & 0.73     \\
 \hline
 U-net ALL\_RC IRM          & 6.7  & 12.0  & 8.5  & 0.76     \\
 U-net ALL\_RC              & {\bf 7.0} & {\bf 13.2} & 8.5 & {\bf 0.79}    \\
 U-net ODD\_RC              & 6.9 & 12.3 & {\bf 8.8} & 0.78    \\
\hline
\end{tabular}
\label{tab:res_babble}
\end{table}


\section{Conclusions}
\label{sec:conclusions}
In this work we proposed U-net-based neural network architectures in which recurrent-convolutional pairs are used at different levels. The obtained result show that the proposed architectures outperform the baseline models (FCLN, RNN, and U-nets without recurrences). The results of the performed experiments suggest, that U-net based architectures perform better for mapping-based rather than masking-based targets.


\bibliographystyle{IEEEbib}
\bibliography{icassp}

\end{document}